\def\year{2021}\relax
\documentclass[letterpaper]{article} 
\usepackage{aaai21}  
\usepackage{times}  
\usepackage{helvet} 
\usepackage{courier}  
\usepackage[hyphens]{url}  
\usepackage{graphicx} 
\usepackage{natbib}  
\usepackage{caption} 
\usepackage{amsmath}
\usepackage{amsfonts}
\usepackage{multirow}
\usepackage{booktabs}

\title{SSD-GAN: Measuring the Realness in the Spatial and Spectral Domains}
\author {
    Yuanqi Chen\textsuperscript{\rm 1,2},
    Ge Li\textsuperscript{\rm 1}\thanks{Corresponding author.},
    Cece Jin\textsuperscript{\rm 1,2},
    Shan Liu\textsuperscript{\rm 4},
    Thomas Li\textsuperscript{\rm 1,3} \\
}
\affiliations {
    \textsuperscript{\rm 1} School of Electronic and Computer Engineering, Peking University \ 
    \textsuperscript{\rm 2} Peng Cheng Laboratory \\
    \textsuperscript{\rm 3} Advanced Institute of Information Technology, Peking University \ 
    \textsuperscript{\rm 4} Tencent America \\
    cyq373@pku.edu.cn, geli@ece.pku.edu.cn, fordacre@pku.edu.cn,
    shanl@tencent.com,
    tli@aiit.org.cn
}

\begin{document}

\maketitle

\begin{abstract}
This paper observes that there is an issue of high frequencies missing in the discriminator of standard GAN, and we reveal it stems from downsampling layers employed in the network architecture. This issue makes the generator lack the incentive from the discriminator to learn high-frequency content of data, resulting in a significant spectrum discrepancy between generated images and real images. Since the Fourier transform is a bijective mapping, we argue that reducing this spectrum discrepancy would boost the performance of GANs. To this end, we introduce SSD-GAN, an enhancement of GANs to alleviate the spectral information loss in the discriminator. Specifically, we propose to embed a frequency-aware classifier into the discriminator to measure the realness of the input in both the spatial and spectral domains. With the enhanced discriminator, the generator of SSD-GAN is encouraged to learn high-frequency content of real data and generate exact details. The proposed method is general and can be easily integrated into most existing GANs framework without excessive cost. The effectiveness of SSD-GAN is validated on various network architectures, objective functions, and datasets. Code will be available
at https://github.com/cyq373/SSD-GAN.
\end{abstract}

\def\EXP{\mathbb{E}}

\section{Introduction}

Generative Adversarial Networks (GANs)~\cite{goodfellow2014generative} involve training a generator and discriminator network in an adversarial manner,
such that the generator learns to reproduce the desired data distribution.
Despite the remarkable achievements in image generation tasks~\cite{isola2017image,karras2019style},
as shown in recent works~\cite{zhang2019detecting,durall2020watch,dzanic2019fourier,frank2020leveraging},
we can efficiently distinguish GAN-generated images from real images in the frequency domain,
which indicates that existing GANs fail to learn the spectral distributions.

\begin{figure}[t]
	\centering
	\includegraphics[width=0.98\linewidth]{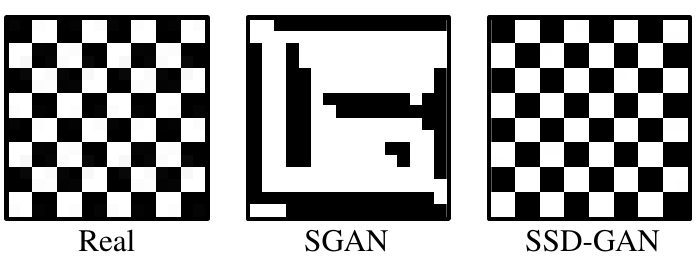}
	\caption{\textbf{Standard GAN (SGAN) fails to learn high frequencies of real data.} 
		In this toy example,
		the real data is a single $16\times16$ image with a checkerboard pattern, which contains significant high frequencies.
		Unlike SGAN,
		our proposed SSD-GAN alleviates the spectral information loss in the discriminator and can well reproduce the real data.
		The detailed experimental setup is described in the supplemental materials.
	}
	\label{fig:toy}
\end{figure}

Recent studies~\cite{dzanic2019fourier,frank2020leveraging} show that the frequency spectrum discrepancy mainly exists at high frequencies.
Because high-frequency components of images influence the exactness of details,
the discrepancy cannot be ignored for generative tasks where details matter.
As shown in Fig.~\ref{fig:toy},
when real data contains significant high frequencies,
standard GAN might fail to reproduce the desired data distribution.
Moreover,
since the Fourier transform is a bijective mapping,
the frequency spectrum discrepancy between real data and the generated samples also indicates that the data distribution in image space is not well captured.
We believe that reducing the spectrum discrepancy would boost the performance of GANs.

In this paper,
we first attempt to explore why there is a spectrum discrepancy between real data and the generated samples.
By investigating the downsampling techniques widely used in the discriminator networks,
we reveal that both of these downsampling strategies,
downsampling with anti-aliasing and downsampling without anti-aliasing,
would lead to high frequencies missing in the discriminator.
Since the training of GANs is a two-player minimax game,
the generator lacks incentives from the discriminator to learn the high-frequency information of the data.

\begin{figure*}[t]
	\centering
	\includegraphics[width=0.98\linewidth]{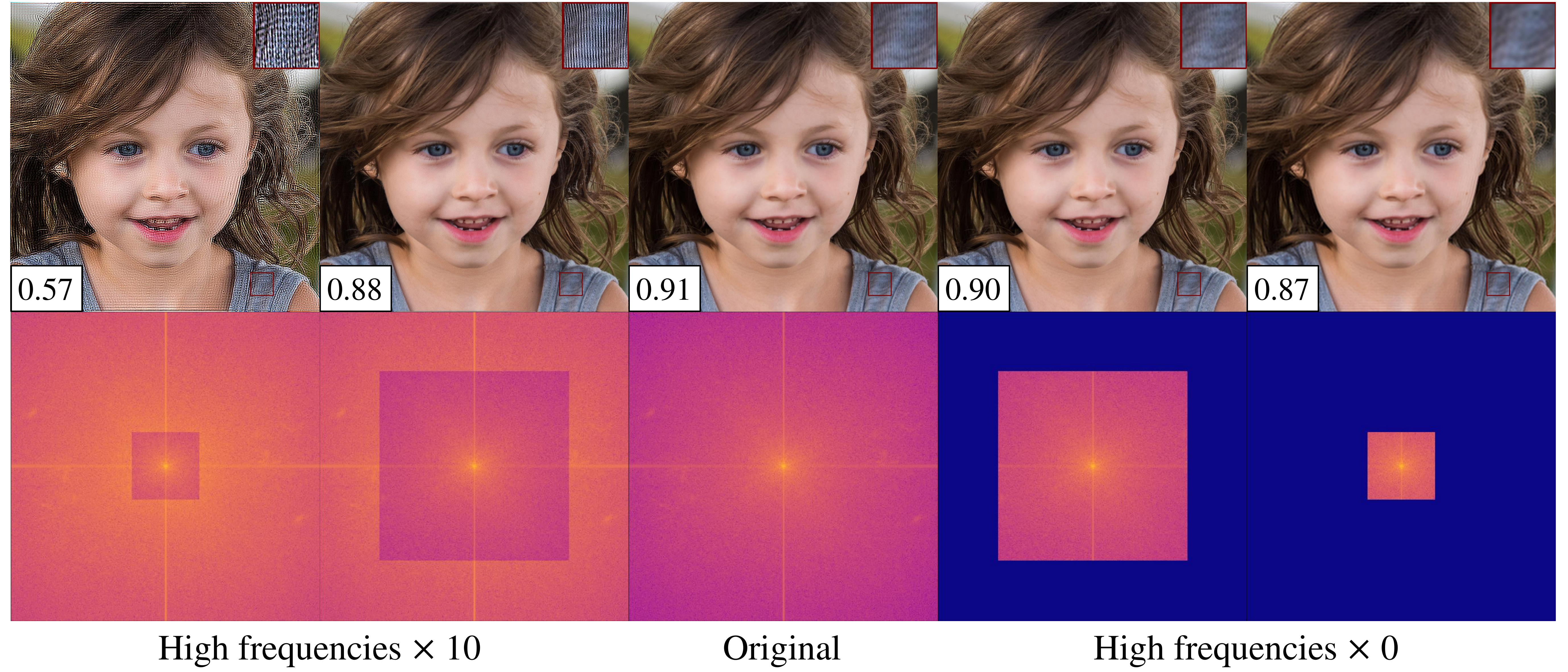}
	\caption{\textbf{The StyleGAN's discriminator cannot distinguish the difference of high frequencies.} 
		The numbers are estimates of $\mathbb{E}[D(x)]$ by averaging 1,000 samples.
		Unless we change the spectrum over a large bandwidth, the outputs of the discriminator are roughly the same,
		which makes StyleGAN fail to reproduce spectral distribution.
	}
	\label{fig:missing}
\end{figure*}

To address the issue of high frequencies missing in the discriminator,
we propose SSD-GAN,
whose discriminator can measure the realness of the input in both the spatial and spectral domains.
To instantiate the idea,
we introduce an additional spectral classifier to detect frequency spectrum discrepancy between real and generated images and integrate it into the discriminator of GANs.
With the enhanced discriminator, 
the generator of SSD-GAN is encouraged to reduce frequency spectrum discrepancy and generate realistic images in both the spatial and spectral domains.
Since a lightweight spectral classifier can be effective,
the proposed method is general and can be easily integrated into most existing GANs framework without excessive cost.
In the experiment,
the effectiveness of the proposed method is validated on various network architectures, objective functions, and datasets.

The contributions of the paper can be summarized as follows:
\begin{itemize}
	\item We observe there is an issue of high frequencies missing in the discriminator of GANs and reveal it stems from downsampling layers employed in the network architecture, which results in a significant spectrum discrepancy between generated images and real images.
	\item We introduce SSD-GAN, an enhancement of GANs to alleviate spectral information loss in the discriminator. 
	With a discriminator that can measure both the spatial and spectral realness of an input sample,
	SSD-GAN can better capture the data distribution than standard GANs.
	\item We show experimentally that the quality of the generations can be improved by reducing the frequency spectrum discrepancy,
	which emphasizes the necessity of learning in the frequency domain.
\end{itemize}

\section{Related Work}

\subsubsection{Generative Adversarial Networks}

Recent rapid advances in Generative Adversarial Networks (GANs)~\cite{goodfellow2014generative}  
have greatly promoted the computer vision and image processing community,
e.g., image inpainting~\cite{yang2018diversity,ren2019structureflow},
image colorization~\cite{isola2017image},
image-to-image translation~\cite{yu2019multi}, etc.
To enhance the quality of generated samples,
PG-GAN~\cite{karras2018progressive} introduces a progressive growing manner for the training process to increase the resolution of synthesized images.
StyleGAN~\cite{karras2019style} propose a style-based generator for finer control over the image synthesis.
Other lines of work focus mainly on improving the discriminator of GANs.
As details are important for generative tasks,
PatchGAN discriminator~\cite{isola2017image} utilizes local discriminator feedback to capture better local structures.
SNGAN~\cite{miyato2018spectral} limits the spectral norm of the weight matrices in the discriminator for Lipschitz constraint.
For countering discriminator forgetting and stabilize the training process,
SS-GAN~\cite{chen2019self} proposes to rotate the image and ask the discriminator to predict the rotation angle.
To effectively balancing the performance of the generator and discriminator,
variational discriminator bottleneck~\cite{peng2018variational} constrains information flow in the discriminator.
In this paper,
after observing the issue of high frequencies missing in the discriminator,
we aim to enhance the ability in the frequency domain of the discriminator.
Thus the generator is encouraged to learn the spectral distribution of real data.

\subsubsection{Frequency Analysis for CNNs}

Even though some GAN-generated images seem to be flawless for human perception,
recent studies~\cite{zhang2019detecting,dzanic2019fourier,frank2020leveraging,durall2020watch} find frequency analysis is effective for image forensics.
They also show that existing GAN based models always fail to reproduce the spectral distribution of real data.
AutoGAN~\cite{zhang2019detecting} first identifies that spectral artifacts stem from upsampling modules included in the GANs pipeline.
To compensate spectral distortions,
a spectral regularization term~\cite{durall2020watch} is proposed to add to the generator loss.
\cite{frank2020leveraging} examines StyleGAN instances using different upsampling techniques and finds bilinear sampling followed by anti-aliasing filters would help to alleviate the problem. 
In this paper,
we investigate another source of spectral distortions, the issue of high frequencies missing in the discriminator.

Apart from frequency analysis for image forensics,
researchers prove that neural networks exhibit a spectral bias~\cite{rahaman2019spectral,xu2019training}; 
they learn filters with a strong bias towards lower frequencies.
Based on this observation,
to effectively control the resource usage,
band-limited convolutional layer~\cite{dziedzic2019band} is introduced to constrain the frequency spectra of filters and data,
while retaining high performance for classification tasks.
However,
high-frequency components cannot be ignored for generative tasks where details matter.
To guarantee all the information can be kept in the model,
MWCNN~\cite{liu2018multi} utilizes discrete wavelet transform (DWT) as a downsampling module in the network architecture for image restoration.
Different from it,
we introduce a spectral classifier to compensate for high-frequency information loss of GANs's discriminator.

\section{High Frequencies Missing
	in the Discriminator}

In the standard GANs~\cite{goodfellow2014generative},
the adversarial loss for the discriminator is defined as:
\begin{equation}
\begin{split}
\mathcal{L}_D =
& -\EXP_{x \sim p_{data}(x)} [\log D(x)] \\
& -\EXP_{x \sim p_g(x)} [\log (1-D(x))],
\end{split}
\end{equation}
and $D(x)$ represents the probability that $x$ comes from $p_{data}$ rather than the generator's distribution $p_g$.
In other words,
$D(x)$ measures the realness of the sample $x$.
If $x$ is realistic,
then it is realistic in all aspects,
such as in the spatial and frequency domains.
However,
as pointed out in recent works~\cite{zhang2019detecting,durall2020watch,dzanic2019fourier,wang2020cnn,frank2020leveraging},
existing GAN based models usually fail to synthesize samples that are realistic in the frequency domain.
It suggests that we cannot only measure the realness in the spatial domain.

Why do these GAN based models fail to reproduce the spectral distributions?
We suspect that the generator lacks incentives from the discriminator to learn the high-frequency information of the data,
since the training of GANs is a two-player minimax game.
To validate the assumption,
we first randomly sample 1,000 images from the real dataset.
Then we modulate the amplitude of high frequencies of different bands.
For the discriminator of StyleGAN~\cite{karras2019style}, 
we compute the mean output of it for images after inverse Fourier transform of the modified spectra.
As shown in Fig.~\ref{fig:missing},
unless we change the spectrum over a large bandwidth,
the discriminator cannot tell the difference of these spectra and the outputs are roughly the same.
As a result,
if the generated images contain some unusual high-frequency components,
the discriminator may not distinguish them to be fake,
which makes StyleGAN fail to reproduce spectral distribution.

\begin{figure}[t]
	\centering
	\includegraphics[width=0.9\linewidth]{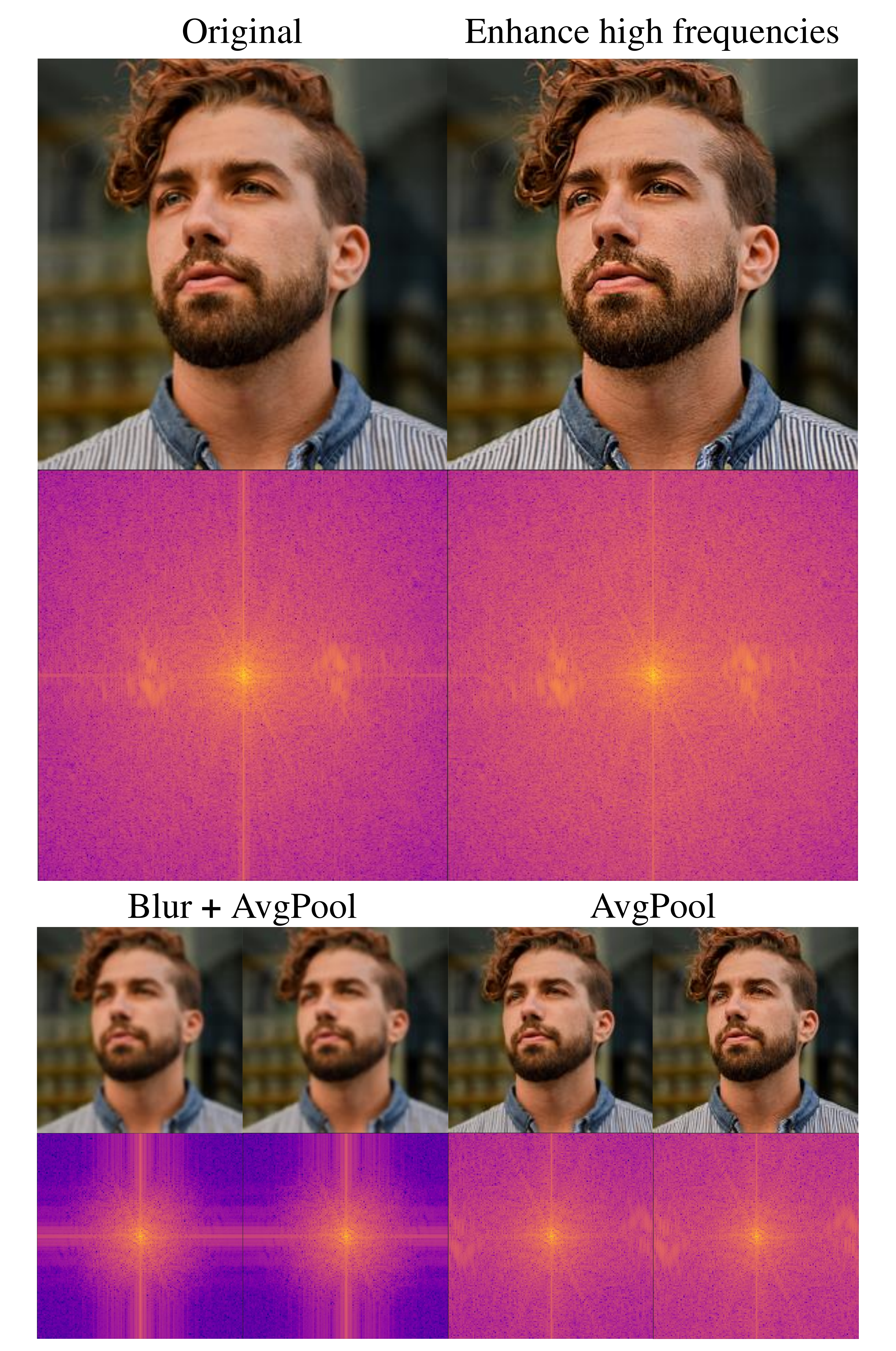}
	\caption{\textbf{Downsampling causes high frequencies missing.} For these two downsampling strategies,
		the left column is the output of the raw image and its spectrum, while the right column is the output of the modulated image and its spectrum. After Blur + AvgPool, there is no significant difference between the spectra before and after the amplitude modulation. For AvgPool, the results exhibit aliasing (e.g., see shirt in the images), which indicates high-frequency details become invalid.
	}
	\label{fig:downsampling}
\end{figure}

Why does the discriminator fail to distinguish the high-frequency contents of the images?
We believe that there is an issue of high frequencies missing in the architecture of the discriminator.
Specifically,
this issue stems from the downsampling modules of the discriminator.
When downsampling an input image,
based on the classical sampling criterion~\cite{nyquist1928certain},
a reasonable approach is to anti-alias by low-pass filtering the image.
Some networks adopt this form of blurred-downsampling~\cite{lecun1990handwritten,karras2019style}.
However,
the low-pass filter removes the high frequencies of the input images.
Since low-pass filtering before downsampling usually results in performance degradation,
it is rarely used today.
Another line of downsampling methods,
such as max-pooling,
strided-convolution, 
and average-pooling,
abandons the use of the low-pass filter. 
However,
these commonly used downsampling methods ignore the sampling theorem~\cite{zhang2019making},
and high-frequency components are aliased and become invalid.
To sum up,
both of these downsampling strategies,
downsampling with anti-aliasing and downsampling without anti-aliasing,
lead to high frequencies missing in the discriminator.

As shown in Fig.~\ref{fig:downsampling},
we provide evidence for the above statement.
For an input image,
we first enhance the amplitude of high frequencies using a sharpening filter.
Then we downsample the raw image and the modulated image and compare the results.
For Gaussian blur followed by average-pooling,
which belongs to downsampling with anti-aliasing,
the high-frequency components are attenuated,
and there is no significant difference between the spectra before and after the amplitude modulation.
For average-pooling,
since it does not provide the anti-aliasing capability,
the results exhibit aliasing (e.g., see shirt in the images), which indicates high-frequency details become invalid.
The more downsampling modules the deep network has,
the wider the bandwidth of the lost high frequencies,
which indicates that the high frequencies missing issue cannot be ignored,
especially for generative tasks where details matter.

\section{Methodology}

In this section,
we first introduce a spectral classifier $C$ to detect frequency spectrum discrepancy between real and generated images.
Then we integrate $C$ into the discriminator of GANs
to enhance its ability in the spectral domain,
thereby reducing the spectrum discrepancy.

\subsection{Detecting Frequency Spectrum Discrepancy}

To address the issue of high frequencies missing in the discriminator,
a straightforward approach is to discriminate in the frequency domain rather than the spatial domain.
For a discrete two-dimensional signal $f(m,n)$ representing an image of size $M \times N$,
we first compute the discrete Fourier transform $\mathcal{F}$ of it,
\begin{equation}
\mathcal{F}(k,l) = \sum_{m=0}^{M-1} \sum_{n=0}^{N-1} f(m,n) e^{- 2 \pi i \frac{km}{M}} e^{- 2 \pi i \frac{ln}{N}},
\end{equation}
for the spectral coordinates $k=0,...,M-1$ and $l=0,...,N-1$.
Then we convert it from Cartesian coordinates $k$ and $l$ to polar coordinates $r$ and $\theta$ for better representing the frequencies of different bands,
\begin{equation}
\begin{split}
\mathcal{F}(r,\theta) = \mathcal{F}(k,l) : r = \sqrt{k^2+l^2}, \theta = atan2(l,k).
\end{split}
\end{equation}
Recent works~\cite{durall2020watch,dzanic2019fourier} have shown that a simple 1D representation of the Fourier power spectrum is effective to highlight the difference between the spectral characteristics of real and deep network generated images.
Following these works,
we get the reduced spectral representation $v$ by azimuthally averaging over $\theta$,
\begin{equation}
\begin{split}
v(r) = \frac{1}{2\pi} \int_0^{2\pi} |\mathcal{F}(r,\theta)| d \theta,
\end{split}
\end{equation}
which represents the mean intensity of the signal with respect to the radial distance $r$.
The reduced spectral representation smooths the fluctuations in the spectrum at high frequencies.

For an input image $x$,
we use the grayscale component of it to get its spectral vector $v$ and denote the process as $v=\phi(x)$.
The spectral classification loss is:
\begin{equation}
\begin{split}
\mathcal{L}_{spectral}~
& = \EXP_{x \sim p_{data}(x)} [\log C(\phi(x))] \\
& + \EXP_{x \sim p_g(x)} [\log (1-C(\phi(x))],
\end{split}
\label{eq:spectral}
\end{equation}
where $C(\phi(x))$ measures the spectral realness of $x$,
and $p_g$ is the generator $G$'s distribution.

\subsection{Reducing Frequency Spectrum Discrepancy}

\begin{figure}[t]
	\centering
	\includegraphics[width=0.95\linewidth]{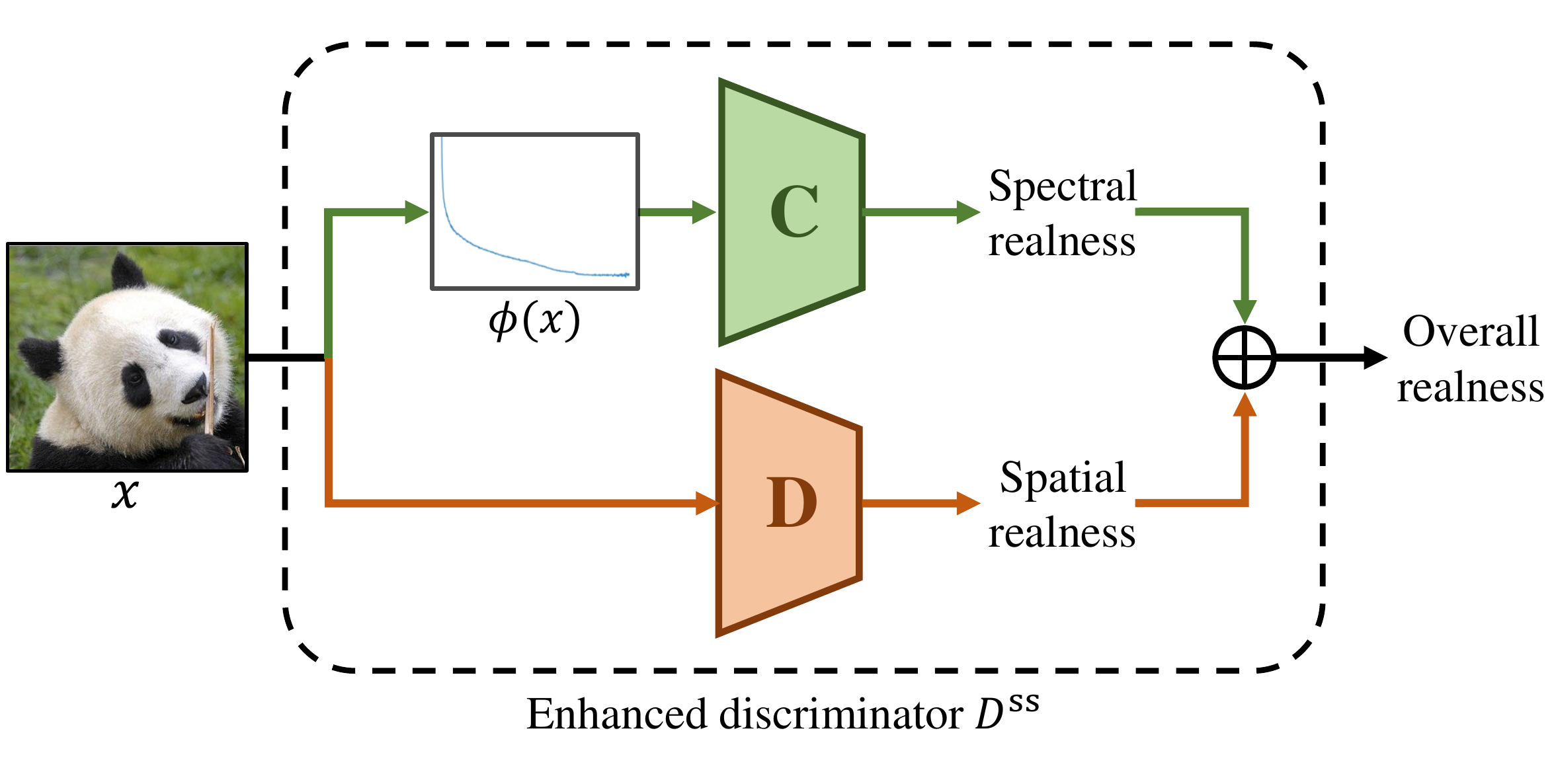}
	\caption{The enhanced discriminator $D^{ss}$ measures both the spectral realness and the spatial realness.
	}
	\label{fig:complemented discriminator}
\end{figure}

Since a sample $x$ is realistic if and only if it is realistic in both the spatial and frequency domains,
we propose to measure the realness of $x$ with the combination of spatial realness and spectral realness.
We integrate the spectral classifier $C$ into the discriminator of GANs to encourage the generator to learn the high-frequency content of the data,
As shown in Fig.~\ref{fig:complemented discriminator},
our enhanced discriminator $D^{ss}$ consists of two modules,
a vanilla discriminator $D$ which measures the spatial realness,
and a spectral classifier $C$.
Thus,
$D^{ss}$ is a discriminator measuring the realness of the input in both the spatial and spectral domains, 
and the overall realness of a sample $x$ is represented as:
\begin{equation}
D^{ss}(x) = \lambda D(x) + (1 - \lambda) C(\phi(x)),
\label{eq:overall realness}
\end{equation}
where $\lambda$ is a hyperparameter that controls the relative importance of the spatial realness and the spectral realness.
The adversarial loss of the framework can be written as:
\begin{equation}
\begin{split}
\mathcal{L}_{adv} ~
& = \EXP_{x \sim p_{data}(x)} [\log D^{ss}(x)] \\
& + \EXP_{x \sim p_g(x)} [\log (1-D^{ss}(x))],
\end{split}
\label{eq:adv}
\end{equation}
where $p_g$ represents the generator $G$'s distribution.

To train our model,
we alternately update spectral classifier $C$,
discriminator $D$,
and generator $G$ with the following gradients:
\begin{equation}
\begin{aligned}
&    \theta_c \leftarrow -\nabla_{\theta_c} \mathcal{L}_{spectral}, \\
&    \theta_d \leftarrow -\nabla_{\theta_d} \mathcal{L}_{adv}, \\
&    \theta_g \leftarrow \nabla_{\theta_g} \mathcal{L}_{adv}.
\end{aligned}
\end{equation}

\subsection{Analyzing the Effect of the Spectral Classifier}

Since much information of the image $x$ is discarded in the spectral vector $\phi(x)$,
we found it cannot provide an effective gradient for 
the adversarial training,
which degrades the performance of the model.
To this end,
we propose that the backpropagation process of Eq.~\ref{eq:adv} does not pass through the spectral classifier $C$,
and $C(\phi(x))$ serves as a spectral modulating factor to the adversarial loss.

\begin{figure}[t]
	\centering
	\includegraphics[width=0.98\linewidth]{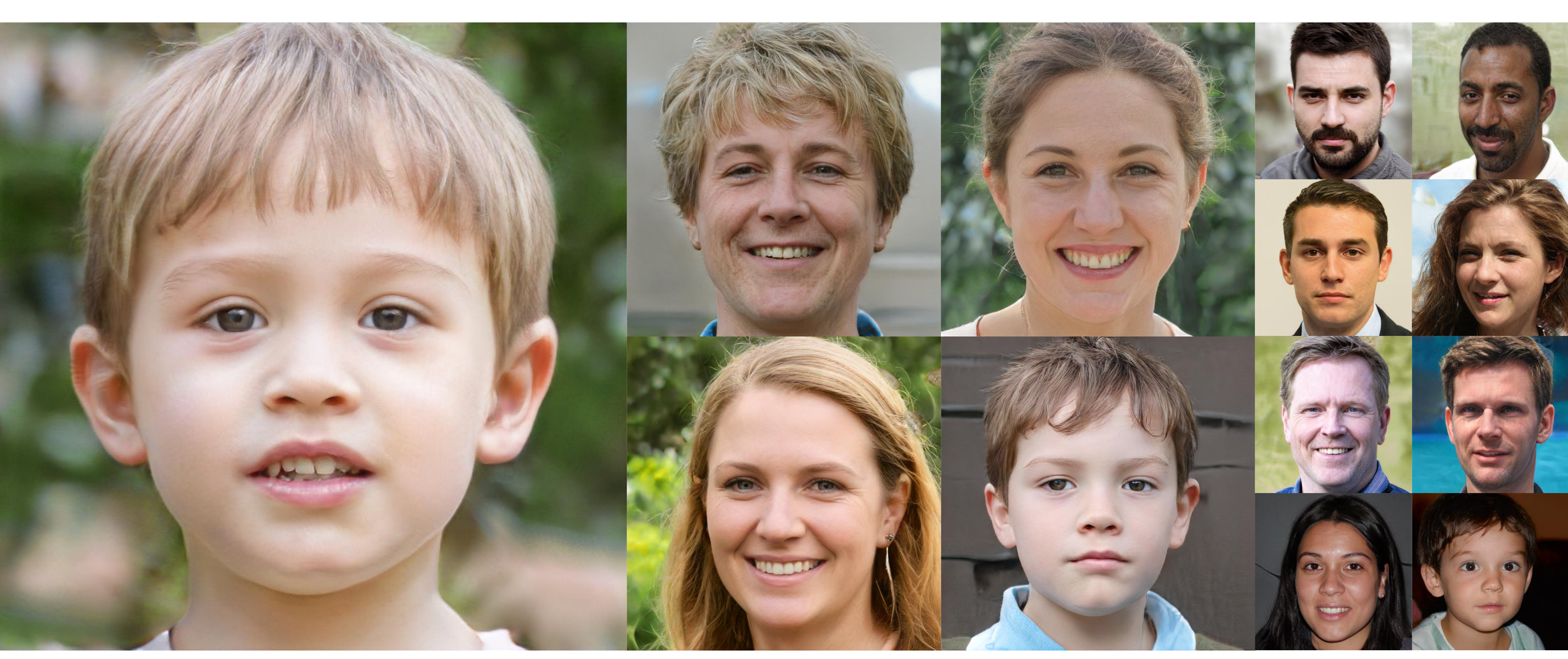}
	\caption{\textbf{FFHQ generations.} Generations of SSD-StyleGAN trained on FFHQ at 1024$\times$1024.
	}
	\label{fig:examples_ffhq}
\end{figure}

We compare the gradient of standard GAN (SGAN) and the proposed method for further insight.
For a generated image $x=G(z)$,
the gradients of the discriminator and generator in non-saturating SGAN are respectively:
\begin{equation}
\frac{1}{1-D(x)}\nabla_{\theta_d} D(x),
\end{equation}
\begin{equation}
-\frac{1}{D(x)}\nabla_{x} D(x) J_{\theta_g} G(z),
\end{equation}
where $\theta_d$ and $\theta_g$ are the parameters of $D$ and $G$, and $J$ is the Jacobian.
As for our method,
the gradients are:
\begin{equation}
\frac{1}{1-D(x)+\frac{1-\lambda}{\lambda}(1-C(\phi(x)))}\nabla_{\theta_d} D(x),
\end{equation}
\begin{equation}
-\frac{1}{D(x)+\frac{1-\lambda}{\lambda}C(\phi(x))}\nabla_{x} D(x) J_{\theta_g} G(z),    
\end{equation}
From these gradients,
it can be observed that our method performs a hard example mining,
where "hard" is defined in the frequency domain.
For the discriminator,
if $C(\phi(x)) \rightarrow 1$,
the generated sample $x$ has good spectral characteristics and is a hard example to be classified as fake.
For the generator,
$x$ is a hard example when $C(\phi(x)) \rightarrow 0$.
This means that $x$ has poor spectral realness and needs more attention from the generator.
In our model,
when $x$ is a hard example in the frequency domain,
the gradients of the discriminator and generator are up-weighted,
which induces the model to learn the spectral distribution of the real data.

\section{Experiments}

Since $\phi(x)$ is easy to compute and $C$ can be a lightweight classifier~\cite{durall2020watch},
the proposed method is general and can be easily integrated into most existing GANs framework without excessive cost.
In the experiment,
we show that for various GANs frameworks of different objective functions, network architectures and datasets,
our method can reduce the frequency spectrum discrepancy and improve the performance in the spatial domain.

\subsection{SSD-StyleGAN}

\subsubsection{Implementation}
Based on StyleGAN~\cite{karras2019style}, 
We evaluate the effectiveness of our method on the FFHQ~\cite{karras2019style} dataset.
It consists 70,000 high-quality images at 1024$\times$1024 resolution.
We use the same implementation as StyleGAN.
In the discriminator,
the activations are blurred before each downsampling layer for anti-aliasing.
The training process is under a progressive growing manner~\cite{karras2018progressive} which starts from 8$\times$8 to 1024$\times$1024.
We apply the non-saturating loss~\cite{goodfellow2014generative} as our adversarial loss with $R_1$ regularization~\cite{mescheder2018training}.
We train all our models with Adam optimizer~\cite{kingma2014adam},
setting $(\beta1, \beta2) = (0, 0.99)$.
The total training time is 25M images.
The hyperparameter $\lambda$ is set to 0.5.
The experiments are conducted on 4 Tesla V100 GPUs.

\begin{figure}[t]
	\centering
	\includegraphics[width=0.98\linewidth]{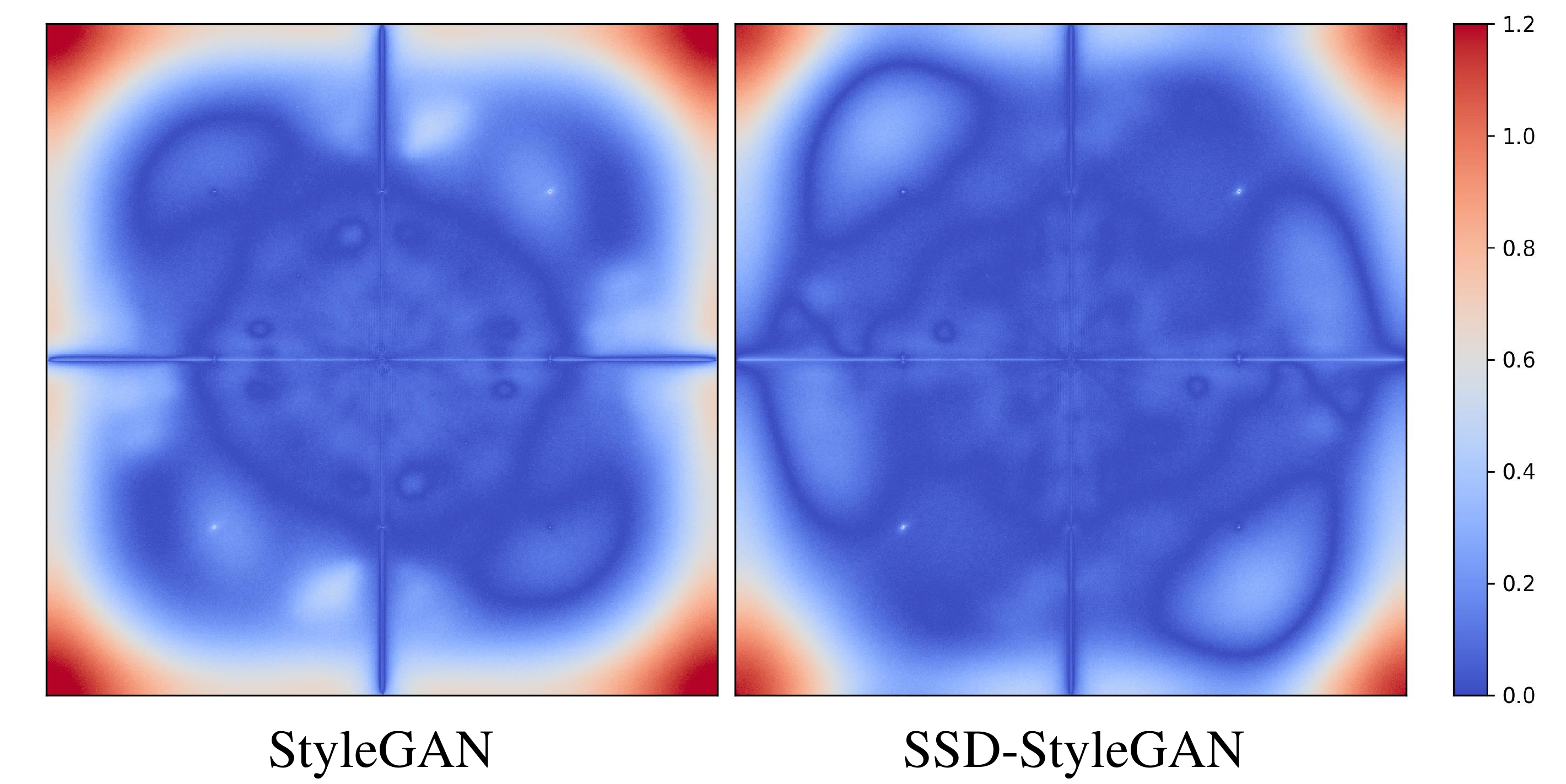}
	\caption{The absolute difference of average spectra $|\EXP [\mathcal{F}(real)] - \EXP [\mathcal{F}(fake)]|$.
	}
	\label{fig:fft_ffhq}
\end{figure}

\subsubsection{Evaluation in the frequency domain}
We first validate whether our proposed SSD-StyleGAN can reduce spectral distortions.
We estimate the average spectrum $\EXP [\mathcal{F}(x)]$ by averaging over 5,000 images.
Then we plot the absolute difference $|\EXP [\mathcal{F}(real)] - \EXP [\mathcal{F}(fake)]|$ between the two average spectra.
As depicted in Fig.~\ref{fig:fft_ffhq},
compared to StyleGAN,
the frequency spectrum discrepancy between images generated by SSD-StyleGAN and real data is significantly reduced.
We also notice that both of the models have spectral distortions at the corners of the spectra that represent extremely high frequencies of images.
In pratice,
due to image compression algorithms applied to real data,
these high-frequency bands contain little information.
Therefore,
to discourage overfitting,
GANs tend not to learn these extremely high frequencies,
and these components of the generated images behave like white noise which has constant power density~\cite{dzanic2019fourier}.

\subsubsection{Evaluation in the spatial domain}
Table~\ref{table:ffhq} reports the performance in the spatial domain.
We adopt Fréchet Inception Distance (FID)~\cite{heusel2017gans} to evaluate the perceptual quality of generated images, and perceptual path length (PPL)~\cite{karras2019style} to measure the degree of disentanglement of representations.
Because the feature extractors used in these metrics are neural networks that map from a high-dimensional input space to a low-dimensional space,
they also suffer from some degree of high-frequency loss and mainly measure the characteristics in the spatial domain.
It is evident that our method performs better than StyleGAN on both these metrics.
We attribute the performance improvement to alleviating the high frequencies missing problem in the discriminator.
By reducing spectral distortions,
it helps to reproduce the spatial distribution of real data,
since the Fourier transform is a bijective mapping.

\begin{figure}[t]
	\centering
	\includegraphics[width=0.98\linewidth]{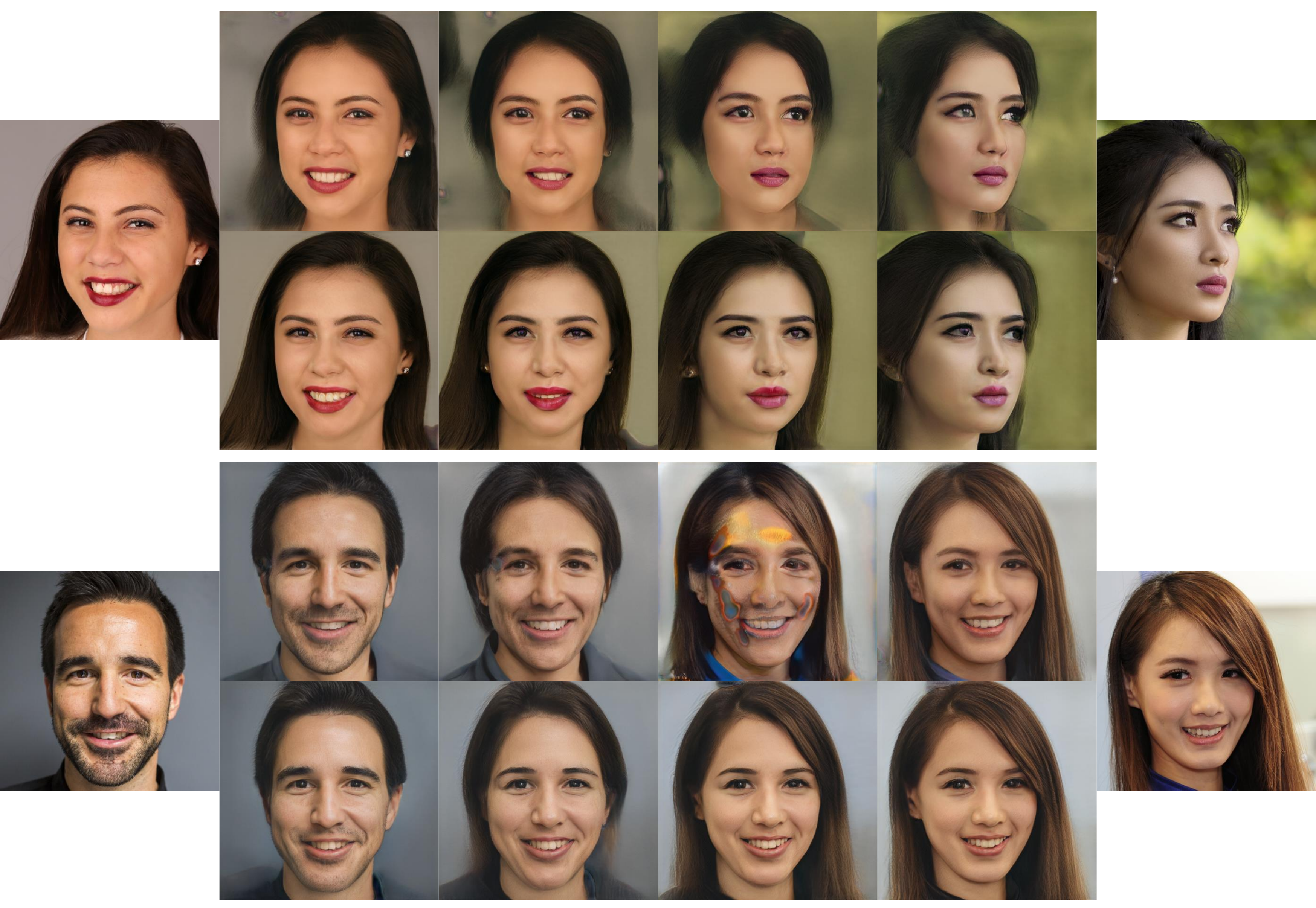}
	\caption{\textbf{Interpolations in the $\mathcal{W}$ space.} For each comparison, the top row represents the reconstructions of the interpolations of StyleGAN, while the bottom row represents the results of SSD-StyleGAN. 
	}
	\label{fig:interpolation}
\end{figure}

For qualitative evaluation,
we utilize the recent embedding algorithm~\cite{abdal2019image2stylegan} to map a given image into the $\mathcal{W}$ space of a pre-trained StyleGAN and then reconstruct back for comparison.
We note that the models may memory images during training and produce good reconstructions.
To avoid overfitting,
we propose to compare the interpolations of StyleGAN and SSD-StyleGAN.
As shown in Fig.~\ref{fig:interpolation},
compared to StyleGAN,
the results of our proposed method show a smoother morphing and have better details,
which is consistent with the quantitative evaluation.
Fig.~\ref{fig:examples_ffhq} shows a collection of generations obtained from SSD-StyleGAN.

\subsubsection{Estimating the spectral quality of samples}

\begin{figure}[t]
	\centering
	\includegraphics[width=0.98\linewidth]{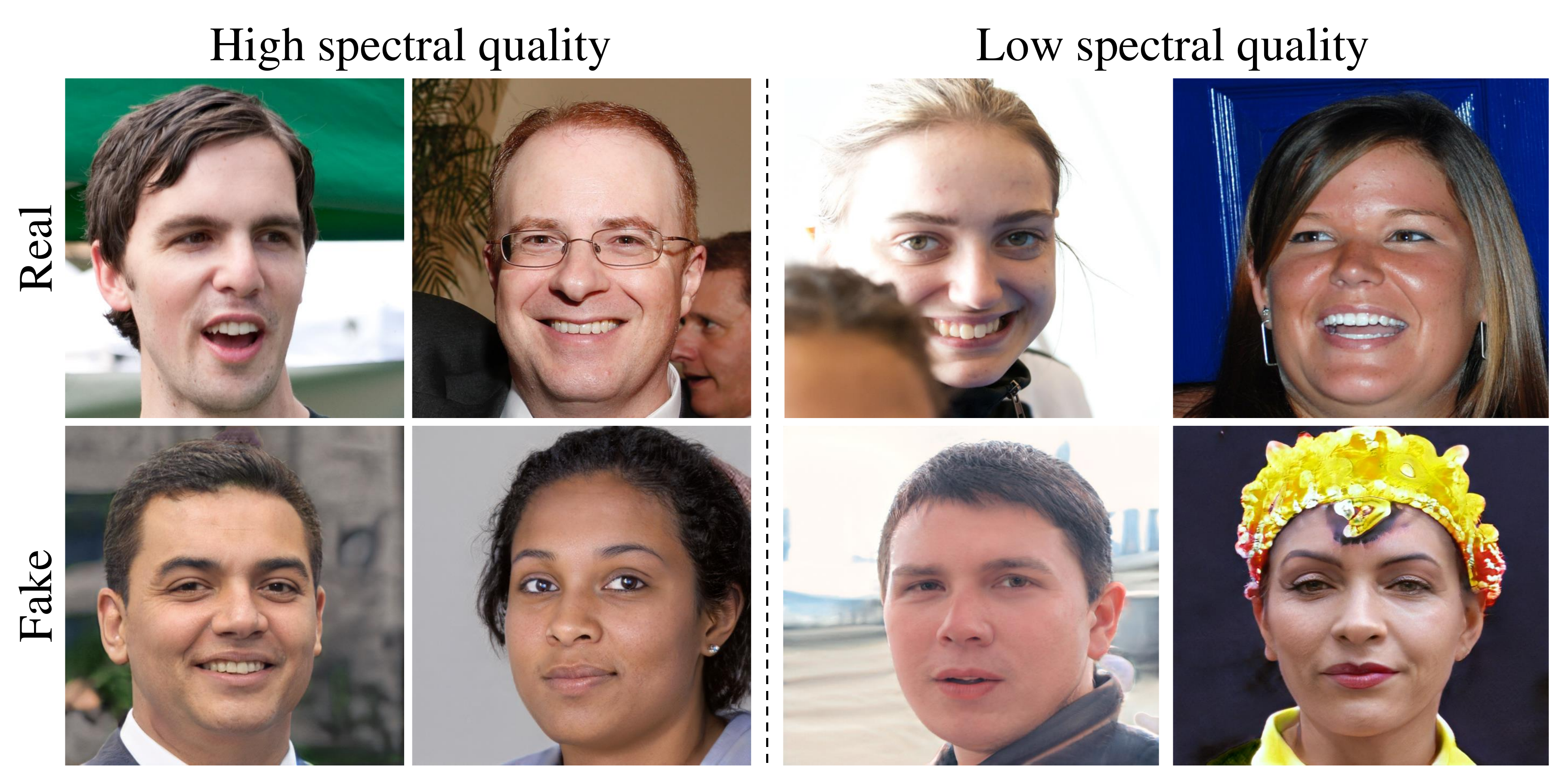}
	\caption{Spectral quality of real data and generations of SSD-StyleGAN.
	}
	\label{fig:test_C}
\end{figure}

In this section,
we evaluate the performance of the spectral classifier $C$ for estimating the spectral quality of samples.
As shown in Fig.~\ref{fig:test_C},
for real data and generations of SSD-StyleGAN,
we present two images with high and low spectral quality scores.
In general,
the samples with high spectral quality scores display a clear portrait.
However,
the images with low spectral quality scores are often overexposed and lose details,
or have some unusual high-frequency components (e.g., see background and headwear in the right column).
The above observation shows the effectiveness of the spectral classifier $C$,
which is beneficial to the learning process of SSD-StyleGAN.

\begin{table}[t]
	\centering
	\begin{tabular}{@{}cccc@{}}
		\toprule
		\multirow{2}{*}{\textbf{Method}} & \multirow{2}{*}{\textbf{FID}} & \multicolumn{2}{r}{\textbf{Path length}} \\ \cmidrule(l){3-4} 
		&                      & \textbf{full}           & \textbf{end}            \\ \midrule
		StyleGAN                & 4.40                 & 234.0          & 195.9          \\
		SSD-StyleGAN                    & \textbf{4.06}                 & \textbf{229.1}          & \textbf{189.7}          \\ \bottomrule
	\end{tabular}
	\caption{FID scores and perceptual path lengths (PPLs) on FFHQ (lower is better). PPLs are measured in the $\mathcal{W}$ space.}
	\label{table:ffhq}
\end{table}

\subsection{SSD-SNGAN}

\subsubsection{Implementation}
Based on SNGAN~\cite{miyato2018spectral},
we evaluate the proposed method on a range of datasets including CIFAR100~\cite{krizhevsky2009learning},
STL10~\cite{coates2011analysis},
and LSUN-bedroom~\cite{yu15lsun}.
Since these datasets have various kinds of resolution,
we mark them with the resolution:
CIFAR100-32,
STL10-48,
and LSUN-128.
We use the same training configurations as ~\cite{lee2020mimicry}.
We train all our models with Adam optimizer~\cite{kingma2014adam},
setting $(\beta1, \beta2) = (0, 0.9)$.
The learning rate is set to 0.0002,
and the minibatch size is 64.
The hyperparameter $\lambda$ is set to 0.5.
All models are trained on a single Tesla V100 GPU.

\subsubsection{Baseline Models} 
We compare our method against three baselines,
including:
\begin{itemize}
	\item \textbf{SNGAN}~\cite{miyato2018spectral} limits the spectral norm of the weight matrices in the discriminator for Lipschitz constraint.
	It adopts a ResNet~\cite{he2016deep} backbone and uses average-pooling as the downsampling layer in the discriminator.
	Different from StyleGAN,
	it utilizes the hinge version\cite{miyato2018spectral} of the adversarial loss.
	\item \textbf{SNGAN+REG}~\cite{durall2020watch} adds a spectral regularization loss to the generator loss to penalize for synthesizing samples with abnormal spectra. The term can be written as $d(\phi(x^{real}),\phi(x^{fake}))$, where $d(\cdot, \cdot)$ measures the binary cross entropy.
	\item \textbf{SNGAN+DWT}~\cite{liu2018multi} adopts discrete wavelet transform (DWT) as downsampling layer to avoid information loss.
	we use 2D Haar wavelet transform to decompose an input into an low-pass representation and three directions of high-requency coefficients. Specifically,
	this DWT downsampling layer transforms the input raw images or a group of feature maps with heith H, width W and channel C into a tensor of shape $(\frac{1}{2} H, \frac{1}{2} W, 4C)$.
\end{itemize}

\begin{figure}[t]
	\centering
	\includegraphics[width=0.95\linewidth]{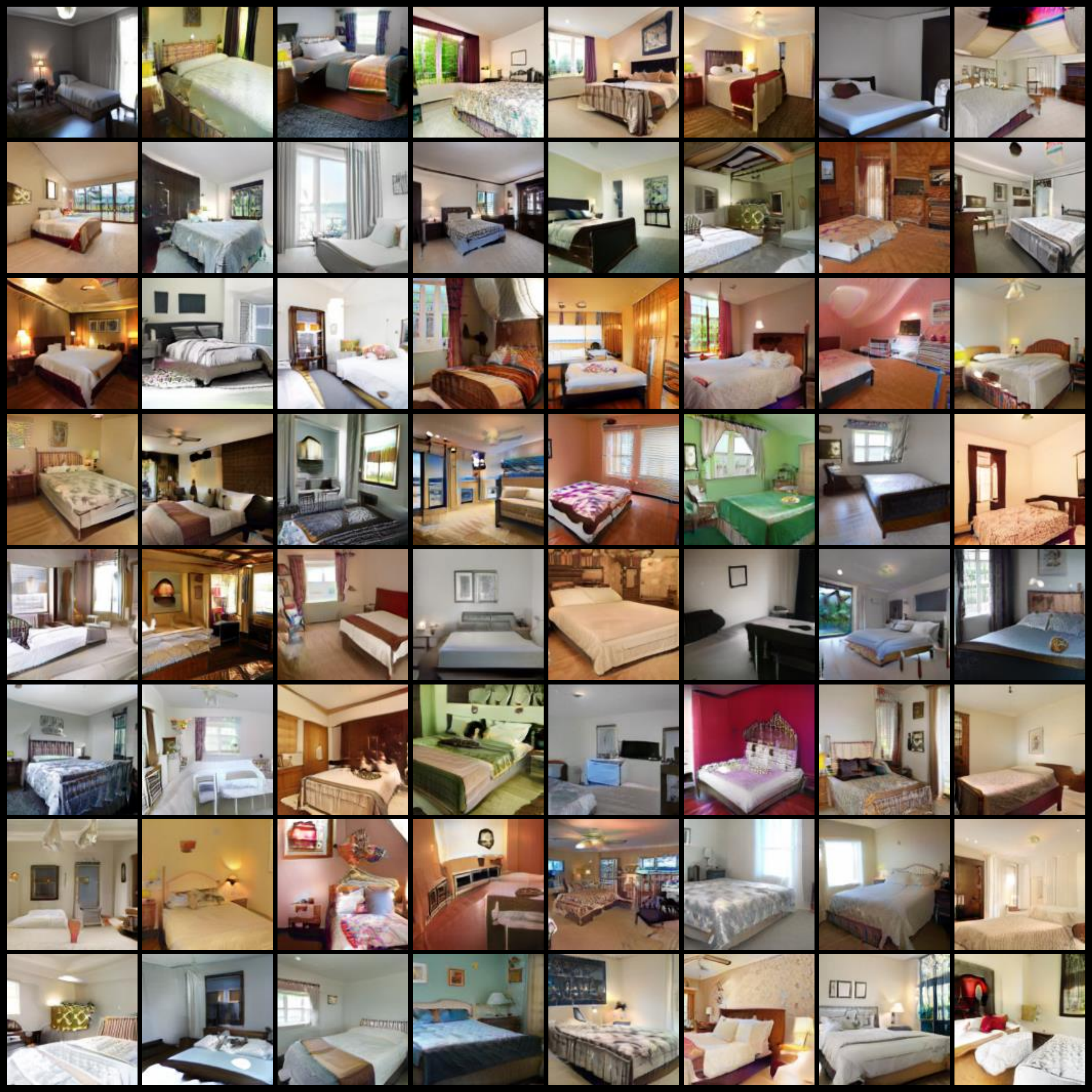}
	\caption{\textbf{Bedroom generations.} Generations of SSD-SNGAN+REG trained on LSUN-bedroom at 128$\times$128.
	}
	\label{fig:examples_lsun}
\end{figure}

\begin{table}[t]
	\centering
	\scalebox{0.85}[0.85]{
		\begin{tabular}{cccc}
			\toprule
			\textbf{Method} & \textbf{CIFAR100-32} & \textbf{STL10-48} & \textbf{LSUN-128} \\ \midrule
			SNGAN           & 22.61                & 39.56             & 25.87             \\
			SNGAN+REG       & 21.39                & 38.16             & 16.95             \\ 
			SNGAN+DWT       & 25.95                & 40.46             & 148.05            \\ \midrule
			SSD-SNGAN       & 19.28                & 36.41             & 15.17             \\
			SSD-SNGAN+REG   & \textbf{19.25}                & \textbf{35.41}             & \textbf{10.61}             \\ \bottomrule
	\end{tabular}}
	\caption{FID scores on CIFAR100-32, STL10-48, and LSUN-128 (lower is better).}
	\label{table:other_dataset}
\end{table}

\subsubsection{Results}

Table~\ref{table:other_dataset} reports the FID scores on CIFAR100-32, STL10-48, and LSUN-128.
SNGAN+REG shows performance improvement over the baseline SNGAN,
indicating that utilizing spectral information is effective.
Compared with SNGAN,
the scores of SNGAN+DWT are higher,
especially for LSUN-128.
We conjecture that because the DWT downsampling layer remains all the high-frequency information of the input,
which contains both details and noises,
it is difficult for the model to learn meaningful semantic representation.
Moreover,
as pointed out by \cite{wang2020high},
learning high-frequency information may degrade the robustness and generalization of a model.
Since LSUN-128 has higher resolution and contains more high-frequency information,
performance degradation of SNGAN+DWT on this dataset is more dramatic.
Our method gets lower FID scores than SNGAN+REG and SNGAN+DWT,
indicating it is a smarter way to utilize high-frequency information of images.
It is remarkable that SSD-SNGAN+REG achieves the best FID scores among all the datasets.
These two techniques have complementary advantages,
since they encourage the model to utilize high-frequency information from the aspects of generator and discriminator respectively. 
Fig.~\ref{fig:examples_lsun} shows a collection of generations obtained from SSD-SNGAN+REG on LSUN-128.

\subsection{Robustness of the Hyperparameter $\lambda$}

\begin{figure}[t]
	\centering
	\includegraphics[width=0.95\linewidth]{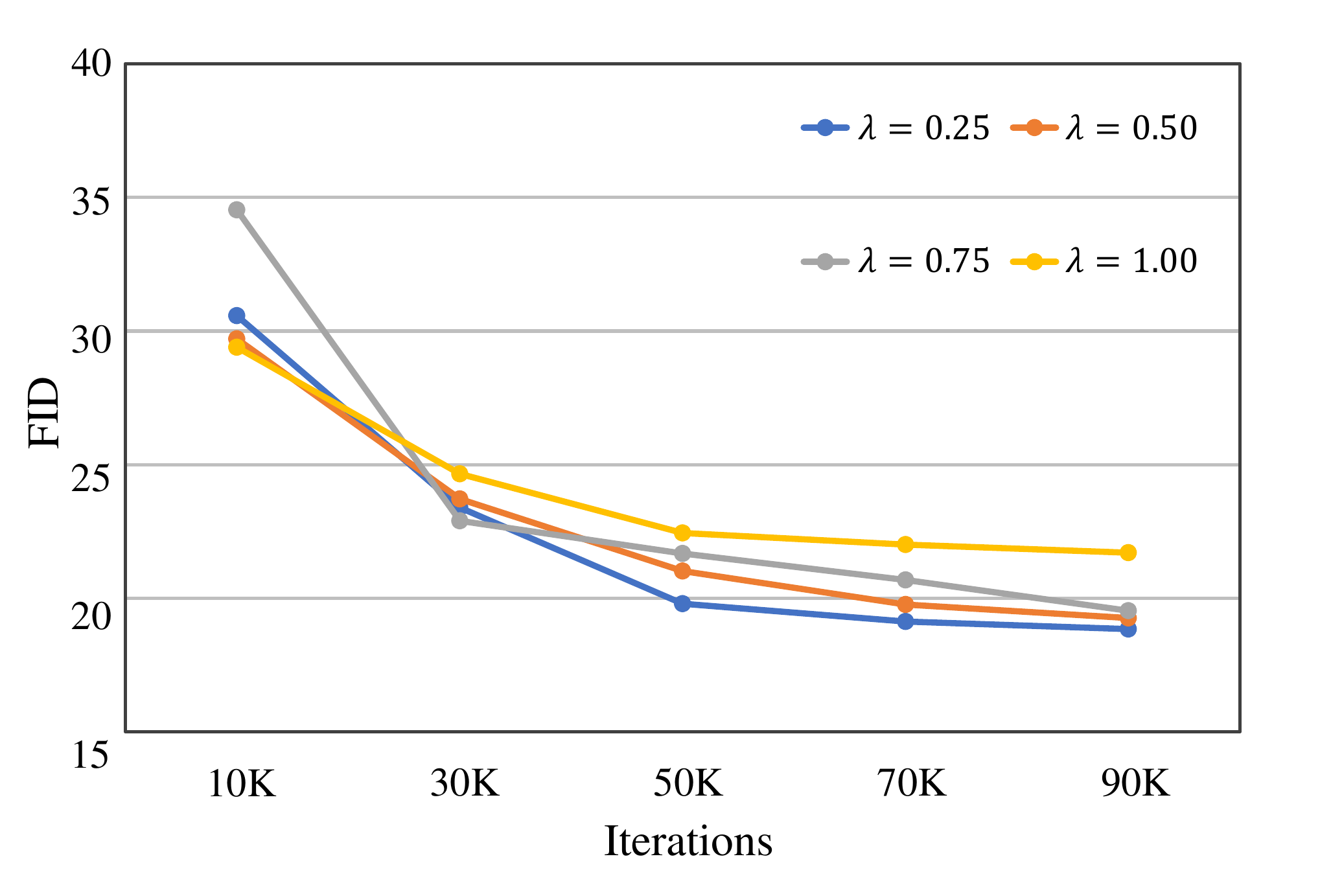}
	\caption{FID scores over the course of training with different hyperparameter $\lambda$.
	}
	\label{fig:lambdas}
\end{figure}

In Eq.~\ref{eq:overall realness},
we introduce a hyperparameter $\lambda$ to control the relative importance of spatial realness and spectral realness.
Here,
we evaluate different hyperparameter settings on CIFAR100-32 to investigate the robustness of $\lambda$.
The baseline model is SNGAN,
and other training settings remain the same as the previous section.
Fig.~\ref{fig:lambdas} compares FID scores over the course of training when setting different values for $\lambda$.
Note that the case of $\lambda=1$ is the baseline model.
We observe that the proposed approach yields consistent performance improvements and enjoys considerable tolerance for the selection of the hyperparameter $\lambda$.

\section{Conclusion and Outlook}

In this paper,
we delve into why existing GANs fail to reproduce the spectral distribution of real data and reveal the issue of high frequencies missing in the discriminator.
To alleviate the issue,
we introduce SSD-GAN,
whose discriminator is enhanced to measure the realness of samples in both the spatial and spectral domains.
We provide empirical evidence that the proposed SSD-GAN can reduce frequency spectrum discrepancy,
thus achieving performance improvement in the image domain.

Frequency analysis provides a novel perspective for analyzing and understanding GANs.
It also opens some avenues for future research.
First,
although recent GAN based models achieve good performance under existing metrics,
the generated samples of these models can easily be distinguished from real data in the frequency domain. 
A metric that quantitatively measure the performance of generative models in the frequency domain would promote the image synthesis community.
Moreover,
besides the discriminator in GANs,
many machine learning tasks involve learning a mapping from a high-dimensional input space to a low-dimensional space.
Constructing a general network architecture to learn semantic representations while taking high frequencies into consideration is interesting and challenging for future work.

\subsubsection{Acknowledgement}
We would like to thank Xiaoming Yu, Hengtao Zhang, Yurui Ren, Nannan Li, and our anonymous reviewers for their valuable feedback and helpful discussions.
This work is supported by National Key R\&D Program of China (No. 2020AAA0103500) and National Natural Science Foundation of China and Guangdong Province Scientific Research on Big Data (No. U1611461).

\bibliography{bib}

\clearpage

\section{Experimental Setup of the Toy Example}

In the toy example,
we aim to describe a simple yet prototypical counterexample to show that standard GAN (SGAN) fails to learn high frequencies of real data.
The real data distribution $p_{data}$ is given by a Dirac-distribution concentrated at a single image,
which has $16\times16$ pixels and a checkerboard pattern with significant high-frequency information.
We train SGAN and SSD-GAN with Adam optimizer,
setting $(\beta_1, \beta_2) = (0,0.9)$.
The learning rate is set to 0.0002.
The models are trained for 10K iterations.
For SSD-GAN,
the hyperparameter $\lambda$ is 0.5.

The network architectures of SGAN and SSD-GAN are shown in Table.~\ref{table:gen} and Table.~\ref{table:dis}.
The spectral classifier $C$ in SSD-GAN has only one fully connected layer.
There are some notations:
N: the number of output channels,
K: kernel size,
S: stride size,
P: padding size,
FC: fully connected layer,
BN: batch normalization,
SN: spectral normalization,
Up: upsampling using bilinear interpolation.

\begin{table}[h]
	\centering
	\scalebox{0.95}[0.95]{
		\begin{tabular}{@{}cc@{}}
			\toprule
			\textbf{Layer}                                                                         & \textbf{Input $\rightarrow$ Output Shape} \\ \midrule
			FC-(8,1024), Reshape                                                                   & (8) $\rightarrow$ (64,4,4)                \\ \midrule
			\begin{tabular}[c]{@{}c@{}}ResBlock: CONV-(N64,K3,S1,P1),\\  BN, ReLU, Up\end{tabular} & (64,4,4) $\rightarrow$ (64,8,8)           \\ \midrule
			\begin{tabular}[c]{@{}c@{}}ResBlock: CONV-(N64,K3,S1,P1),\\  BN, ReLU, Up\end{tabular} & (64,8,8) $\rightarrow$ (64,16,16)         \\ \midrule
			\begin{tabular}[c]{@{}c@{}}BN, ReLU,\\ CONV-(N1,K3,S1,P1), Tanh\end{tabular}                                                     & (64,16,16) $\rightarrow$ (1,16,16)        \\ \bottomrule
	\end{tabular}}
	\caption{Architecture of the generator $G$.}
	\label{table:gen}
\end{table}

\begin{table}[h]
	\centering
	\scalebox{0.95}[0.95]{
		\begin{tabular}{@{}cc@{}}
			\toprule
			\textbf{Layer}                                                                                   & \textbf{Input $\rightarrow$ Output Shape} \\ \midrule
			\begin{tabular}[c]{@{}c@{}}ResBlock: CONV-(N128,K3,S1,P1), \\ ReLU, AvgPool-(K2,S2)\end{tabular} & (1,16,16) $\rightarrow$ (128,8,8)         \\ \midrule
			\begin{tabular}[c]{@{}c@{}}ResBlock: CONV-(N128,K3,S1,P1), \\ ReLU, AvgPool-(K2,S2)\end{tabular} & (128,8,8) $\rightarrow$ (128,4,4)         \\ \midrule
			\begin{tabular}[c]{@{}c@{}}ResBlock: CONV-(N128,K3,S1,P1), \\ ReLU\end{tabular}                  & (128,4,4) $\rightarrow$ (128,4,4)         \\ \midrule
			ReLU, GlobalSumPool                                                                              & (128,4,4) $\rightarrow$ (128)             \\ \midrule
			FC-(128,1), SN                                                                                     & (128) $\rightarrow$ (1)                   \\ \bottomrule
	\end{tabular}}
	\caption{Architecture of the discriminator $D$.}
	\label{table:dis}
\end{table}

\section{Additional Qualitative Results}

\begin{figure}[h]
	\centering
	\includegraphics[width=0.85\linewidth]{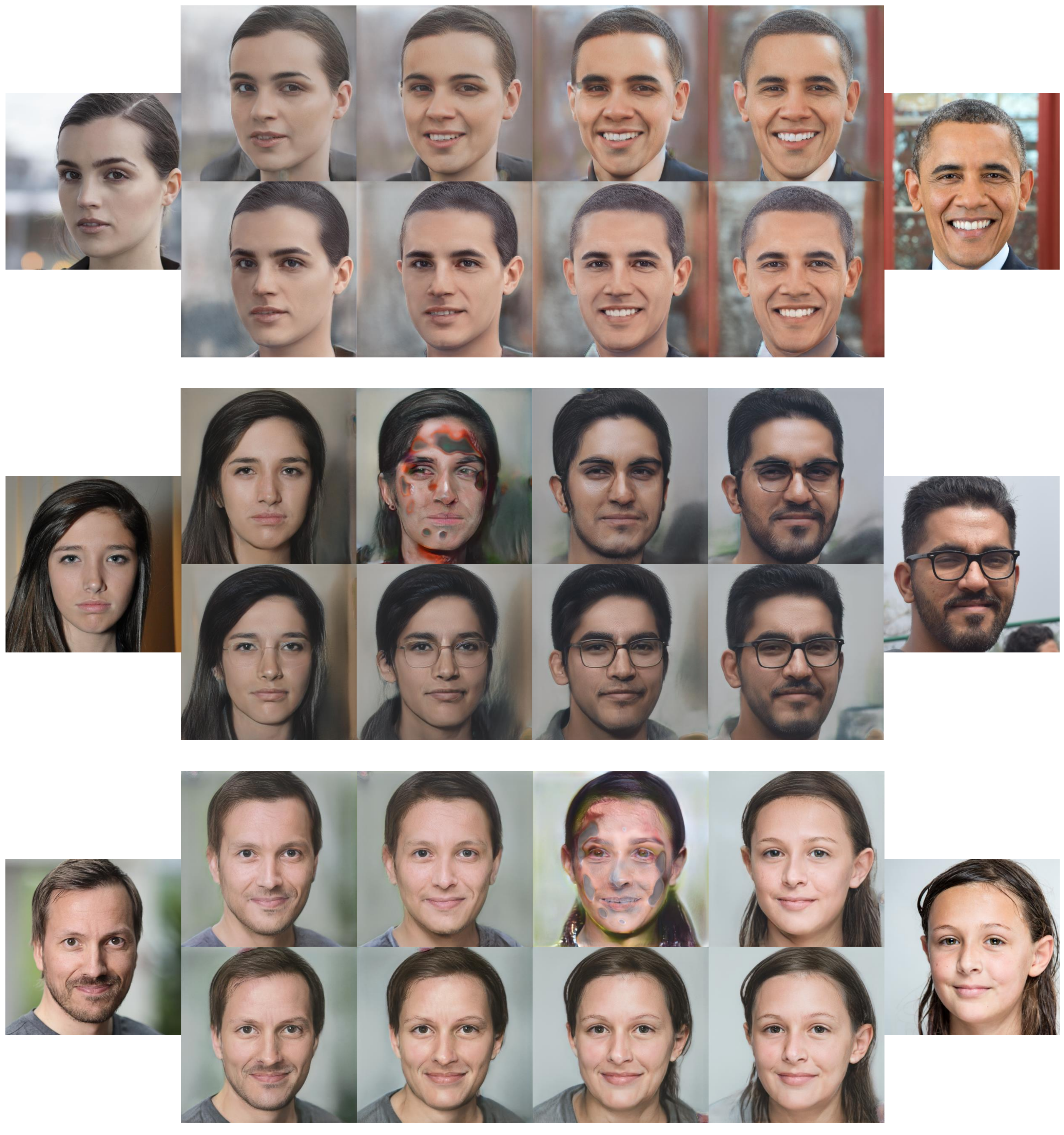}
	\caption{\textbf{Interpolations in the $\mathcal{W}$ space.} For each comparison, the top row represents the reconstructions of the interpolations of StyleGAN, while the bottom row represents the results of SSD-StyleGAN. 
	}
	\label{fig:interpolation_supp}
\end{figure}

\begin{figure}[h]
	\centering
	\includegraphics[width=0.9\linewidth]{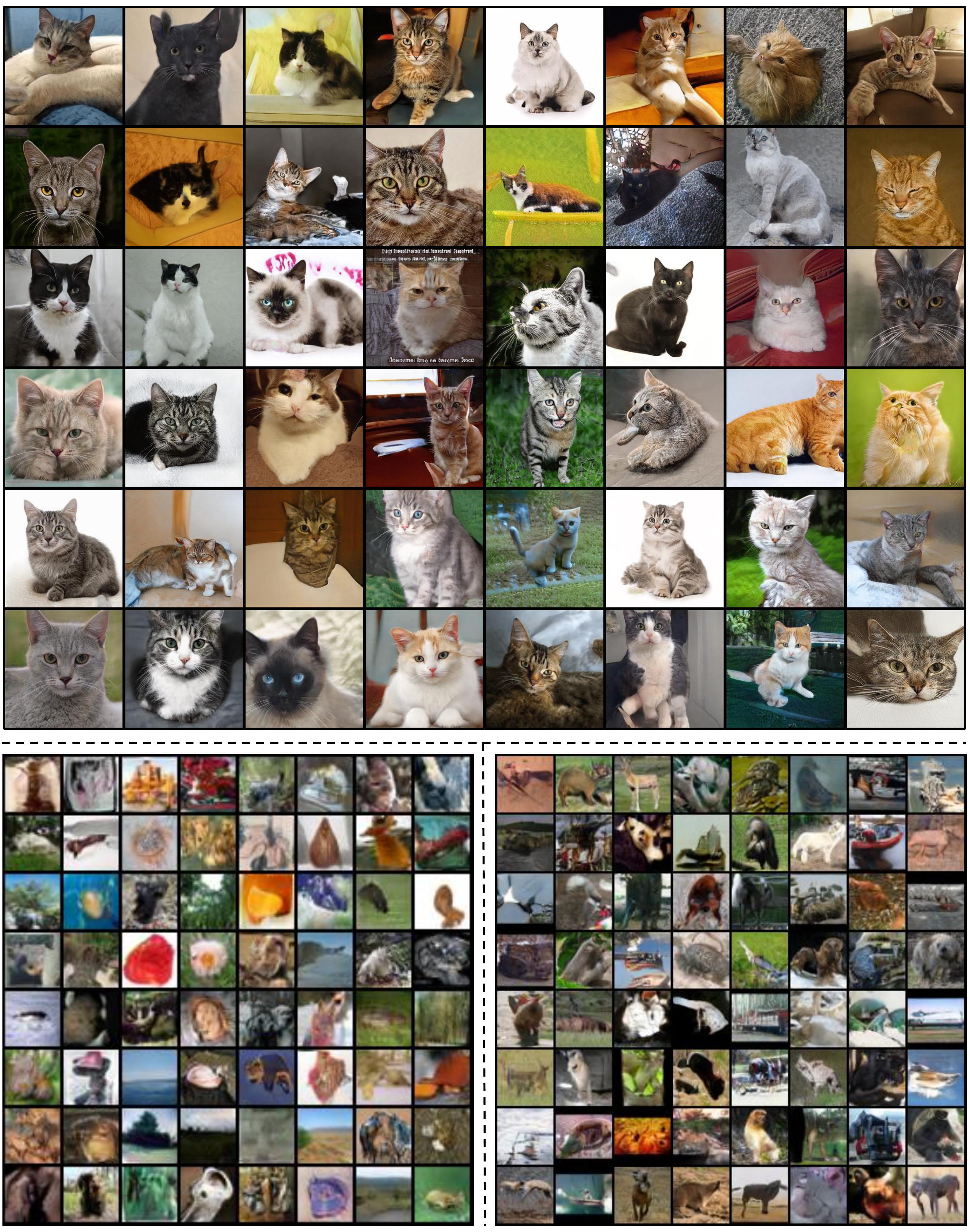}
	\caption{\textbf{Generations of multiple datasets.} Top: Generations of SSD-StyleGAN trained on LSUN-CAT dataset.
	Bottom-left: Generations of SSD-SNGAN trained on CIFAR100 dataset.
	Bottom-right: Generations of SSD-SNGAN trained on STL-10 dataset.
	}
	\label{fig:examples_supp}
\end{figure}

We provide more qualitative results for interpolations in Fig.\ref{fig:interpolation_supp} and generations of multiple datasets in Fig.\ref{fig:examples_supp}.
As shown in Fig.\ref{fig:interpolation_supp},
compared to StyleGAN, the results of our proposed method show a smoother morphing and have better details.
Fig.\ref{fig:examples_supp} shows more generated samples of our proposed method trained on multiple datasets including LSUN-CAT, CIFAR100 and STL-10.

\end{document}